\documentclass[10pt,twocolumn,letterpaper]{article}
\usepackage[pagenumbers]{cvpr}

\definecolor{cvprblue}{rgb}{0.21,0.49,0.74}
\usepackage[breaklinks,colorlinks,allcolors=cvprblue]{hyperref}
\title{UnrealPose: Leveraging Game Engine Kinematics for Large-Scale Synthetic Human Pose Data}
\author{}
\begin{document}
\twocolumn[{%
\renewcommand\twocolumn[1][]{#1}%
\begin{center}
{\Large\bfseries UnrealPose: Leveraging Game Engine Kinematics for Large-Scale Synthetic Human Pose Data\par}
\vspace{8pt}
{\normalsize
Joshua Kawaguchi \quad
Saad Manzur \quad
Emily Gao Wang \quad
Maitreyi Sinha \quad
Bryan Vela \quad
Yunxi Wang \quad
Brandon Vela \quad
Wayne B.~Hayes \par}
\vspace{4pt}
{\normalsize University of California, Irvine\par}
\vspace{2pt}
{\ttfamily\small
\{jkawagu1, smanzur, egwang1, maitres, bjvela, yunxiw5, bovela, whayes\}@uci.edu\par}
\end{center}
\vspace{10pt}
\begin{center}
\captionsetup{type=figure}
\includegraphics[width=\textwidth]{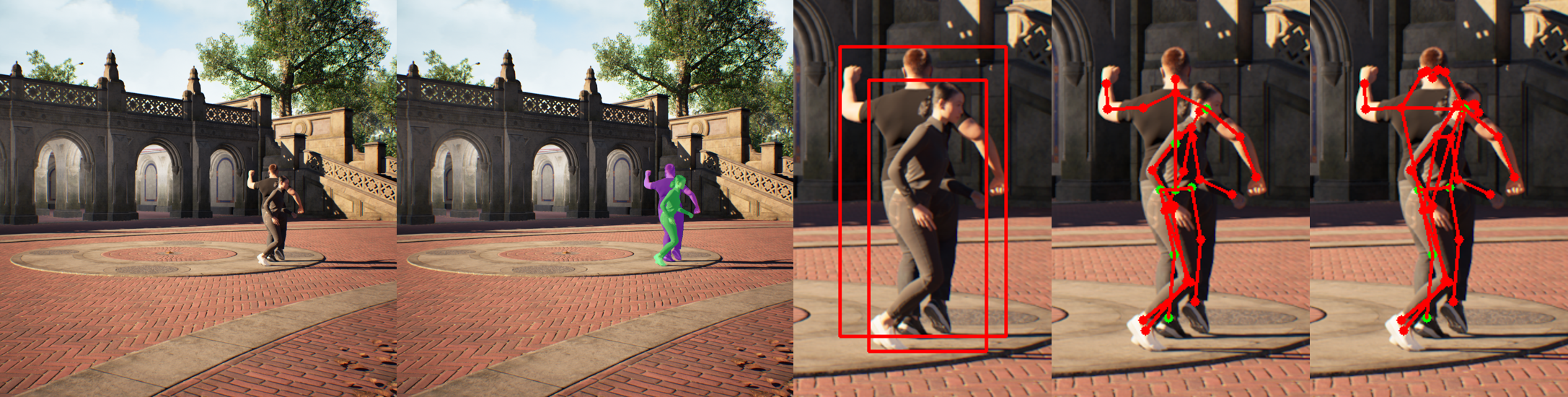}
\captionof{figure}{Visualization of annotations in UnrealPose-1M. The first two panels show the full-resolution RGB frame and its instance mask overlay. The remaining panels show cropped regions for easier viewing, containing person bounding boxes, standard 2D keypoints, and COCO keypoints with occlusion flags (green).}
\label{fig:main}
\end{center}
}]
\begin{abstract}
Diverse, accurately labeled 3D human pose data is expensive and studio-bound, while in-the-wild sets lack known ground truth. We introduce UnrealPose-Gen, an {\it Unreal Engine 5} pipeline built on {\it Movie Render Queue} for high-quality offline rendering. Our generated frames include: (i) 3D joints in world and camera coordinates, (ii) 2D projections and COCO-style keypoints with occlusion and joint-visibility flags, (iii) person bounding boxes, and (iv) camera intrinsics and extrinsics. We use UnrealPose-Gen to present UnrealPose-1M, an approximately one million frame corpus comprising eight sequences: five scripted “coherent” sequences spanning five scenes, approximately 40 actions, and five subjects; and three randomized sequences across three scenes, approximately 100 actions, and five subjects, all captured from diverse camera trajectories for broad viewpoint coverage. As a fidelity check, we report real-to-synthetic results on four tasks: image-to-3D pose, 2D keypoint detection, 2D-to-3D lifting, and person detection/segmentation. Though time and resources constrain us from an unlimited dataset, we release the UnrealPose-1M dataset, as well as the UnrealPose-Gen pipeline to support 3rd-party generation of human pose data.
\end{abstract}
\section{Introduction}
\label{sec:intro}

Learning accurate human pose from images depends on large datasets with reliable supervision. Controlled-3D datasets provide precise annotations but are constrained in action diversity, viewpoints, and backgrounds ~\cite{h36m_pami, DBLP:journals/corr/JooSLLTGBGNMKNS16, mono-3dhp2017, TotalCapture}, limiting generalization. In contrast, in-the-wild datasets offer diversity at scale, but usually only provide hand-labeled 2D keypoints ~\cite{lin2015microsoft, 6909866, li2019crowdposeefficientcrowdedscenes} that are prone to inter-annotator error ~\cite{DBLP:journals/corr/JooSLLTGBGNMKNS16}, or require specialized capture rigs that restrict locations and camera motion to obtain 3D keypoints ~\cite{vonMarcard2018}.

Because accurate real-world 3D is hard to obtain, many works derive pseudo 3D supervision from abundant 2D data. Two methods are primarily used for this task: lifting models that predict 3D joints from 2D joints ~\cite{h36m_pami, mde, 9010332, 9710217, Zhao_2019}, and parametric models that regress SMPL or other human body models from images ~\cite{bogo2016smplautomaticestimation3d, kolotouros2019learningreconstruct3dhuman, kanazawa2018endtoendrecoveryhumanshape, kocabas2021pareattentionregressor3d, kocabas2022specseeingpeoplewild}. The accuracy of lifting models trained in one dataset generally loses significant precision when applied to different datasets ~\cite{manzur2025posebench3dcrossdatasetanalysisframework, gholami2022adaptposecrossdatasetadaptation3d, 10.1007/978-3-031-73007-8_18} with different cameras, subjects, or poses, reflecting the under-constrained nature of 2D-to-3D lifting. Parametric models inherit fundamental biases from SMPL's training data (the CAESAR ~\cite{robinette2002caesar} dataset of 18-65 year old subjects from Italy, Netherlands, and North America), limiting representation of diverse body types and special cases like amputees. Furthermore, fitting methods introduce additional biases, e.g., some approaches producing bent knees ~\cite{dwivedi_cvpr2024_tokenhmr} while others generate unnaturally straight legs ~\cite{tiwari2022posendfmodelinghumanpose, weng2022domainadaptive3dpose}. Their joints, derived through regression rather than anatomical modeling, accumulate errors from these cascading biases, resulting in pseudo ground truths that are unreliable.

The limitations of pseudo-3D supervision have driven researchers toward synthetic data generation, which promises perfect ground truth annotations and scalable generation across diverse scenes, subjects, and motions. Recent advances have shown strong human pose and shape results with SMPL-based supervision. However, most existing synthetic datasets remain mesh-centric, primarily targeting mesh recovery from images, with 2D and 3D pose reconstruction as secondary outputs. Pose-centric pipelines require more precise joint supervision, explicit occlusion information, and occlusion-aware person detection.

Ironically, while the computer vision community struggles to synthetically generate complex human interactions---with even state-of-the-art datasets acknowledging that ``generating realistic synthetic sequences of general human-object and human-human interaction remains an open research problem'' ~\cite{tesch2025bedlam2}---these exact scenarios are abundant in video game animation libraries. Game developers have spent decades creating complex multi-person interaction, object manipulation, and diverse character movements. We propose to tap the rich video game heritage of native kinematics to produce a fundamentally novel engine for generating human pose datasets. Unlike mesh-centric datasets ~\cite{black2023bedlamsyntheticdatasetbodies, tesch2025bedlam2, varol17_surreal, patel2021agoraavatarsgeographyoptimized} that rely on SMPL models, our human animations and labels are based directly on Unreal Engine (UE) MetaHumans and their skeletal joints---the pivot points that drive character animation in the engine. This SMPL-independent approach eliminates the SMPL biases and enables us to leverage the full ecosystem of UE-compatible content: marketplace assets, retargeted MoCap, and---crucially---the interaction-rich animations common in games (such as multi-person fighting, conversations, and object interactions). While we use MetaHumans for our dataset, any UE skeleton and animation can be rendered with full annotations, allowing the computer vision community to leverage existing, sophisticated game animation assets rather than trying to synthetically recreate what game developers have already perfected.

To this end, we introduce {\it UnrealPose-Gen}: a UE5/Movie Render Queue (MRQ) generator producing high-quality images with corresponding annotations. Each frame includes: (i) kinematic 3D rig joints in both world and camera coordinates, (ii) COCO-style ~\cite{lin2015microsoft} 2D keypoints with per-joint visibility, (iii) person bounding boxes and segmentation masks, and (iv) complete camera intrinsics/extrinsics. Beyond offline rendering via MRQ, our camera pipeline supports online rendering during gameplay, enabling real-time dataset generation from UE5 games or applications. This opens unprecedented opportunities for domain-specific data generation using existing game content.

We leverage UnrealPose-Gen to generate—and release—{\it UnrealPose-1M}: a dataset of one million annotated frames with diverse subjects, actions, and environments. To validate the fidelity and realism of our synthetic data, we evaluate on four fundamental tasks: person instance segmentation, 2D keypoint detection, 2D → 3D lifting, and image → 3D pose regression. We report  real-to-synthetic results to measure how closely our data matches real-world distributions to demonstrate practical utility. Both the generator and dataset are publicly released to facilitate future research.

\section{Related Work}
\label{sec:related_work}
\subsection{Human Pose Datasets}

\noindent\textbf{Real-World 3D Pose Datasets.} 
Early 3D pose datasets relied heavily on controlled laboratory settings with marker-based motion capture systems. Human3.6M ~\cite{h36m_pami}, MPI-INF-3DHP ~\cite{mono-3dhp2017}, and 3DPW ~\cite{vonMarcard2018} are some of the most widely used benchmarks. However, these datasets remain limited in scale due to the cost and complexity of 3D annotation equipment and they lack the interaction-rich scenarios common in real-world applications ~\cite{eva, gpa}.

\noindent\textbf{2D Pose Datasets.}
2D datasets such as COCO-Pose ~\cite{lin2015microsoft}, MPII ~\cite{6909866}, and CrowdPose ~\cite{li2019crowdposeefficientcrowdedscenes} provide extensive in-the-wild annotations through manual labeling. While these datasets offer diversity in poses, appearances, and environments, they lack the 3D annotations which are important for many applications. 

Furthermore, manual annotations suffer from human variability. Ronchi et al. ~\cite{DBLP:journals/corr/RonchiP17} explicitly model this inconsistency through Object Keypoint Similarity (OKS). Similar pixel level disagreement has been documented in other domains such as the Partially Occluded Hands dataset~\cite{hands}, which reports per-keypoint inter-annotation mean pixel error for joints both visible and occluded. While human annotation noise is inevitable, synthetic data can provide perfect ground truth.

\subsection{Learning 3D from 2D}
\noindent\textbf{2D-to-3D Lifting}
A common approach to obtaining 3D supervision involves lifting 2D keypoints to 3D space. Most methods now use machine learning to map 2D tracks to 3D, often with temporal models (e.g., VideoPose3D ~\cite{vpose}). However, these methods suffer from poor generalization. SOTA lifting networks can achieve 30-50mm Mean Per Joint Position Error (MPJPE) on Human3.6M but they also suffer huge cross dataset performance drops ~\cite{manzur2025posebench3dcrossdatasetanalysisframework, gholami2022adaptposecrossdatasetadaptation3d, 10.1007/978-3-031-73007-8_18}. This limitation motivates the need for more reliable 3D supervision.

\noindent\textbf{Parametric Model Fitting}
SMPL and its successors (SMPL-x, STAR, SUPR, SKEL, etc.) ~\cite{SMPL:2015, MANO:SIGGRAPHASIA:2017, SMPL-X:2019, STAR:2020, SUPR:2022, keller2023skel} provide a parametric representation of human body shape and pose. Fitting methods such as SMPLify ~\cite{bogo2016smplautomaticestimation3d}, SPIN ~\cite{kolotouros2019learningreconstruct3dhuman}, and HMR ~\cite{kanazawa2018endtoendrecoveryhumanshape} estimate these parameters from optimization or model fitting in the loop. The resulting 3D “joints” are produced by applying a joint regressor to the fitted mesh ~\cite{SMPL:2015}, so their locations depend on both the fit and chosen regressor rather than the kinematic pivots. Moreover, SMPL family models do not natively support hair, deformations, contain spurious long-range correlations ~\cite{black2023bedlamsyntheticdatasetbodies}, and their shape space reflects the demographics of the scans used to train them(e.g., CAESAR ~\cite{robinette2002caesar}). Consequently, labels derived from fitting parametric models inherit method and model induced biases.

\subsection{Synthetic Human Pose Datasets}

\noindent\textbf{SMPL-based synthetic corpora.}
SURREAL pioneered synthetic supervision by rendering SMPL sequences driven by motion capture and compositing them over background images; however, it uses simplified textures and shading and requires fine-tuning on real images for competitive performance.
AGORA ~\cite{patel2021agoraavatarsgeographyoptimized} fits SMPL-X models to high-quality scans and renders photoreal multi-person scenes for mesh-centric evaluation; it is image-centric rather than sequence-centric. 

BEDLAM ~\cite{black2023bedlamsyntheticdatasetbodies} advances photoreal SMPL-X supervision in monocular RGB videos with diverse bodies, clothing, and hair, and reports that models trained purely on synthetic data can reach state-of-the-art human pose-and-shape (HPS) on real images; BEDLAM-2 ~\cite{tesch2025bedlam2} further expands motions and coverage.

\noindent\textbf{Limitations of SMPL-centric supervision.}
Despite these advances, most synthetic resources are designed for mesh recovery. Labels and metrics center on SMPL parameters, while joint-level pose is typically derived by applying a learned joint regressor to the mesh. This means (i) joint locations depend on both the fitted mesh and the chosen regressor rather than kinematic rotation centers, (ii) the fixed minimally-clothed topology limits native modeling of loose clothing, hair, and complex contact/collision without additional mechanisms, and (iii) demographic coverage reflects the scans used to build the models (e.g., CAESAR), and (iv) complex interactions remain an ""open research problem"" as acknowledged by BEDLAM's authors ~\cite{tesch2025bedlam2}. These choices are ideal for HPS, but less aligned with pose pipelines that require calibrated 2D-3D pairs, explicit per-joint visibility, and occlusion-aware detection labels.

\noindent\textbf{Engine-native labels (ours).}
In contrast, we label engine-native kinematic joints from the UE skeletal pivots that drive animation. Our generator is SMPL-independent meaning any permitted UE skeleton and animation (marketplace packs, mocap retargets, even SMPL motions retargeted to UE rigs) can be rendered with per-frame intrinsics/extrinsics, COCO-style 2D keypoints with per-joint visibility, and occlusion-aware person masks/boxes. This exposes interaction-rich motions (e.g., combat sports, collaborative tasks, tool manipulation) that are standard in game libraries but difficult to motion-capture safely.

\subsection{Summary}
Prior work either offers accurate but constrained controlled-3D capture, large-scale 2D-only supervision with human variability, lifting and parametric fits that import method/model biases, or mesh-centric synthetic corpora that prioritize SMPL mesh recovery. We instead provide pose-centric supervision: engine-native kinematic joints from UE skeletons, calibrated 2D-3D keypoint pairs with per-joint visibility and occlusion-aware boxes/masks, and a reusable UE5 generator that can tap into decades of professional game animation assets.

\section{Methods}
\label{sec:methods}

We present UnrealPose-Gen, a UE5 pipeline for generating annotated human pose data, and UnrealPose-1M, a one million frame dataset demonstrating its capabilities. Our approach leverages engine-native skeletons to produce calibrated 2D→3D pose pairs with explicit occlusion information. 
\subsection{UnrealPose-Gen}
\noindent\textbf{Camera-Centric Architecture.}
UnrealPose-Gen is built entirely within the camera system, enabling both real-time online rendering during gameplay and offline rendering via MRQ. This unified architecture ensures consistent annotation quality regardless of rendering mode. Users select which character assets to track and the system extracts annotations from the camera's perspective for all tracked characters.

\noindent\textbf{Character Tracking System.}
We allow the users to select up to 255 character assets for tracking and annotation export. Once selected, the system monitors these characters, maintaining the instance IDs and generating per-frame annotations. This approach works with MetaHumans, any custom characters, or any UE-compatible skeletal mesh, provided it has a valid skeleton.

\noindent\textbf{Camera Configuration.}
Our system supports arbitrary camera parameters. Users can set any focal length, sensor size, aspect ratio, etc., and the pipeline will function properly. Since our current implementation uses static cameras, we export the intrinsic/extrinsic matrix once per camera rather than per frame.

\noindent\textbf{Resolution and Format Independence.}
Users can render at any resolution and the annotation system will scale accordingly, projecting 3D joints to the correct pixel coordinates for the chosen output resolution. Export formats are user-configurable, supporting various image formats (PNG, JPG, EXR).

\noindent\textbf{Annotation Generation.}
For each rendered frame, the system annotates:
\begin{enumerate}
    \item \textbf{3D Joint Positions:} The system queries the skeletal mesh component to obtain the world-space coordinates of all specified joints in each tracked character's skeleton. We then transform these coordinates to camera space. These positions represent the pivot points that drive animations. 
    \item \textbf{2D Keypoint Projection:} We provide two separate sets of 2D keypoint annotations: the 3D joints projected into 2D image coordinates and the standard COCO-Pose keypoints.
    \item \textbf{Per-Joint Visibility Flags:} For all keypoints, we perform a line trace from the camera to the world point to determine occlusion.
    \item \textbf{Occlusion-Aware Person Detection Labels:} We generate bounding boxes and instance segmentation masks for each tracked character. Both bounding boxes and segmentation masks are occlusion aware. If a person is partially occluded by objects or other people, the segmentation mask is cut off at the occluding boundary, and the bounding box tightly bounds around only the visible portion of each person. Each person is assigned a unique instance ID that is maintained across frames to identify them in the mask.
\end{enumerate}
While we export certain common keypoints for the generation of UnrealPose-1M, the pipeline is easily customizable. Users can modify a few lines of code to export any subset of joints from the skeleton, including eyes, ears, individual hand joints, facial landmarks, or any other skeletal joint available in their character rig.

\noindent\textbf{Data Filtering.}
To ensure data quality and reduce redundancy, we apply two filtering criteria before saving frames:
\begin{enumerate}
    \item \textbf{Frame Boundaries:} We discard frames where a tracked person has keypoints that project outside the image bounds. This ensures that all saved frames contain subjects fully in frame.
    \item \textbf{Temporal Redundancy:} Per tracked person, we compare the Euclidean distance of joints and discard frames that are not sufficiently different. This filtering reduces near duplicate frames while maintaining coverage, improving dataset efficiency.
\end{enumerate}

\noindent\textbf{Online Rendering Support.}
While UnrealPose-1M was generated using MRQ to ensure the highest quality, UnrealPose-Gen supports real-time online rendering during gameplay. This capability opens significant opportunities for domain-specific generation where researchers could generate training data directly from existing UE5 games. The annotation system operates in the same way, ensuring consistency between online and offline generation. We view this capability as a major benefit to our approach, enabling future work in real-time data generation.

\subsection{UnrealPose-1M}
\noindent\textbf{Character Movement Driver.}
To generate UnrealPose-1M, we developed a movement driver that supports two modes: scripted movements and random movements. This allows us to balance temporal coherence with diverse coverage.

\noindent\textbf{Scripted Movement Mode.}
In scripted mode, we define specific markers. Characters move between markers while playing specified locomotion animations. While stationary, an appropriate idle animation is selected as well. This mode produces coherent sequences suitable for video-based pose estimation methods that rely on temporal consistency.

\noindent\textbf{Random Movement Mode.}
In random mode, we define an area for characters to explore and provide a directory full of animations. The system randomly selects a location within the bounded area and selects random animations to play while moving and idle. This mode maximizes diversity in poses, viewpoints, and actions, producing data suitable for single frame methods and improving generalization. 

\noindent\textbf{Coherent Sequences.} Using scripted movement mode, we created five coherent sequences:
\begin{itemize}
    \item \textbf{Subjects:} Five distinct MetaHumans with varied body types, skin tones, hairstyle, and clothing.
    \item \textbf{Actions:} Approximately 40 scripted actions including locomotion (walking, running), stationary poses (sitting, standing), and environment-specific interactions that form coherent animation sequences.
    \item \textbf{Environments:} Five distinct scenes across two UE5 levels—an art gallery and a basketball court—with different locations within each level providing environmental diversity.
    \item \textbf{Camera Coverage:} Each sequence uses 15-20 static camera positions with varying FOV and degrees of elevation to capture characters.
\end{itemize}

\noindent\textbf{Randomized Sequences.} Using random movement mode, we created three randomized sequences:
\begin{itemize}
    \item \textbf{Subjects:} The same five MetaHumans used in coherent sequences, providing appearance consistency across dataset splits.
    \item \textbf{Actions:} A library of approximately 100 animations sampled during generations, including complex motions, extreme poses, and interactions.
    \item \textbf{Environments:} Three distinct scenes within a city park level, each offering vastly different visual characteristics and spatial layouts.
    \item \textbf{Camera Coverage:} Each sequence uses multiple static camera positions with varying FOV and degrees of elevation to capture characters.
\end{itemize}

\begin{figure}[t]
\centering
\includegraphics[width=0.9\linewidth]{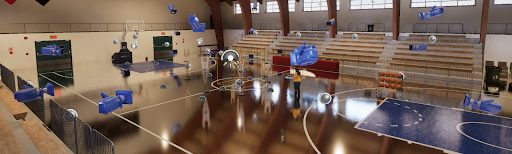}
\caption{Example camera configuration. We vary FOV (30°--90°), camera height (ground to overhead), and distance to subject, producing unconventional viewpoints including ground-level and steep overhead angles rarely present in existing datasets.}
\label{fig:camera_setup}
\end{figure}

\noindent\textbf{Camera Configuration.} Across both sequence types, we  varied camera parameters to maximize coverage:
\begin{itemize}
    \item \textbf{FOV Range:} We varied FOV from approximately 30$^{\circ}$ to approximately 90$^{\circ}$ to capture a wide variety of camera configurations.
    \item \textbf{Camera Heights:} Camera heights were varied from ground level to overhead positions.
    \item \textbf{Distance to Subject:} Since our cameras are static, each sequence produces frames where the subject appears at both close and far distances.
\end{itemize}
This variation in camera configuration exposes models to viewpoints that are rarely present in standard benchmarks, which are typically collected from canonical third-person views with limited camera pose diversity.

\noindent\textbf{Annotations Per Frame.}
Each frame provides: (i) 17 COCO-Pose format 2D keypoints with visibility flags, (ii) 16 common skeletal joints projected to 2D with visibility flags, (iii), 16 common 3D joints in world and camera coordinates, (iv) per-person bounding boxes and segmentation masks with unique IDs.

\noindent\textbf{Dataset Composition.}
UnrealPose-1M comprises approximately one million annotated frames. We provide a 75/20/5 train/validation/test split of frames that are at least 100mm Euclidean distance apart (sum of all joints in camera coordinates) from the previous frame in sequence.
\begin{itemize}
    \item \textbf{Coherent sequences:} Approximately 800,000 frames across five scenes
    \item \textbf{Randomized sequences:} Approximately 170,000 frames across three scenes.
    \item \textbf{Multi-person frames:} Approximately 115,000 across two scenes.
\end{itemize}

\section{Experiments}
Because complete training on the UnrealPose-1M dataset was not feasible under our compute constraints, our experiments focus on a comprehensive evaluation study rather than training new models from scratch. We benchmark multiple models directly on our synthetic data to measure: (i) real-to-synthetic transfer, (ii) The fidelity of UnrealPose-1M, and (iii) the impact of multi-person interactions, occlusions, and scene diversity on model performance. This allows us to validate data quality and realism without access to the computational resources needed for full model training.

We evaluate on our test set of 38,050 randomly sampled frames for 2D-to-3D lifting, image-to-3D regression, and instance segmentation. Our sample spans all eight environments and both coherent and randomized sequences to ensure proper coverage. Additionally, we evaluate image-to-2D keypoint detection on a smaller subset.


\begin{table*}[t]
\centering

\begin{tabular}{lcccccc}
\toprule
\textbf{Model} &
\textbf{AP} & \textbf{AP50} & \textbf{AP75} & \textbf{AR} &
\textbf{MPJPE (mm)} & \textbf{PA-MPJPE (mm)} \\
\midrule
HRNet-W48 (2D) & 0.883 & 0.990 & 0.980 & 0.896 & -- & -- \\
DEKR-HRNet-W32 (2D) & 0.802 & 0.977 & 0.923 & 0.831 & -- & -- \\
PoseAug (2D$\rightarrow$3D) & -- & -- & -- & -- & 61.81 & 57.28 \\
MeTRAbs (Img$\rightarrow$3D) & -- & -- & -- & -- & 104.16 & 111.41 \\
\bottomrule
\end{tabular}

\vspace{6pt}
\caption{\textbf{Quantitative evaluation on UnrealPose-1M.}
(a) AP, AP$^{50}$, AP$^{75}$, and AR evaluate Image$\rightarrow$2D keypoint detection using COCO metrics.
Top-down HRNet-W48 is evaluated on the full 2D test split comprising 38,050 images and 42,528 annotated person instances, while bottom-up DEKR-HRNet-W32 is evaluated at the image level on the same 38,050-frame test set.
(b) MPJPE and PA-MPJPE evaluate 2D$\rightarrow$3D lifting and Image$\rightarrow$3D pose regression.
High 2D keypoint scores reflect annotation consistency and metric compatibility rather than task difficulty.}
\label{tab:quant}
\end{table*}

\label{sec:experiments}
\subsection{Image → 2D Keypoint Detection}
UnrealPose-1M is designed to support standard COCO-style 2D keypoint evaluation. 
Each frame provides the 17 COCO keypoints with per-joint visibility flags and 
engine-native person bounding boxes~\cite{lin2015microsoft}, making the dataset 
compatible with the top-down and bottom-up paradigms widely used in human pose 
estimation~\cite{sun2019hrnet, geng2021dekr, cheng2020higherhrnet}. 
In this section, we outline the intended evaluation protocol and report small-scale 
sanity checks that validate our annotation and conversion pipeline.

For 2D keypoint detection, we construct a held-out test split spanning all eight
sequences and a wide range of scenes, camera viewpoints, subject counts, and
occlusion patterns. The resulting test set contains 38,050 images with a total
of 42,528 annotated person instances, reflecting the presence of multiple people
in many frames. Following the COCO keypoint evaluation protocol~\cite{lin2015microsoft},
we compute Object Keypoint Similarity (OKS) and report Average Precision (AP),
AP$^{50}$, AP$^{75}$, and Average Recall (AR).

Because UnrealPose-1M includes multi-person interactions and substantial occlusion, we additionally recommend 
reporting performance stratified by (i) the number of people per frame and 
(ii) the fraction of occluded joints, enabling explicit analysis of robustness 
to crowding and occlusion.

UnrealPose-1M naturally supports two complementary evaluation regimes:

\paragraph{Top-down.}
For top-down detectors such as HRNet~\cite{sun2019hrnet}, oracle person crops can 
be extracted directly from our engine-native bounding boxes. This isolates the 
pose estimation module from person detection, enabling controlled studies on camera 
pose, subject diversity, and occlusion.

\paragraph{Bottom-up.}
Bottom-up methods such as DEKR~\cite{geng2021dekr} and HigherHRNet~\cite{cheng2020higherhrnet} 
can be evaluated on full-resolution frames using our instance IDs and segmentation 
masks to assess keypoint grouping accuracy and robustness in multi-person, 
interaction-heavy scenes.

To verify the correctness of our COCO conversion and evaluation pipeline, we 
conducted preliminary tests using publicly released COCO-pretrained models from 
MMPose~\cite{mmpose2020} on the full 2D test split. A top-down HRNet model~\cite{sun2019hrnet} and a bottom-up DEKR 
model~\cite{geng2021dekr} both achieve strong but non-saturated AP and AR scores which is consistent with expectations for COCO-trained models evaluated on a synthetic domain. These results confirm that the evaluation protocol is fully compatible with standard COCO metrics while still leaving room for future domain-adapted training and model improvements.

A comprehensive benchmark across additional architectures and training regimes is left for future work.

\subsection{2D → 3D Lifting}
\begin{figure*}[t]
    \centering
    \includegraphics[width=\textwidth]{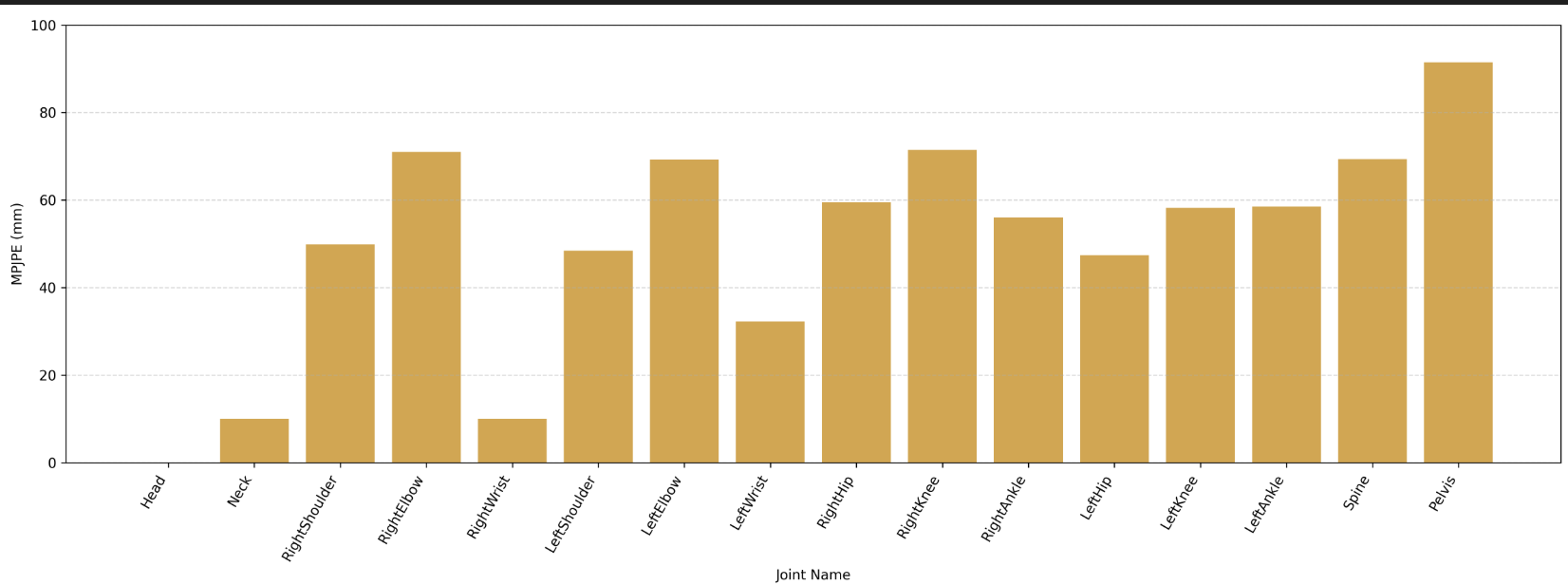}
    \caption{Per-joint MPJPE distribution across 16 body joints on evaluation of PoseAug.}
    \label{fig:per_joint_mpjpe}
\end{figure*}

To evaluate how well our synthetic data supports lifting, we use PoseAug~\cite{poseaug}, a widely adopted cross-domain 2D-to-3D regression model, as an evaluator. PoseAug is originally trained on Human3.6M~\cite{h36m} and has never been fine-tuned on our synthetic dataset. Therefore, it provides an independent and meaningful metric for examining the geometric consistency between the rendered images and their 3D annotations. We sampled approximately 55,000 synthetic frames as evaluation samples to obtain stable statistical behavior. For each frame, we predicted 16 3D joints using the publicly available PoseAug model and performed root-joint translation alignment between predictions and ground truth to ensure that the evaluation focuses on the relative skeletal geometry rather than its global position. As shown in Table~\ref{tab:quant}, PoseAug achieves a Mean Per Joint Position Error (MPJPE) of 61.813 mm and Procrustes-Aligned MPJPE(PA-MPJPE) of 57.277 mm. Given that the lifting model is trained entirely on real data, these values fall within the expected range reported in prior synthetic-to-real cross-dataset studies, indicating strong geometric consistency between our 2D keypoints and the corresponding 3D annotations.

As shown in Fig.~\ref{fig:per_joint_mpjpe}, the per-joint MPJPE distribution reveals meaningful anatomical patterns. Torso related joints such as the neck, spine, and hip joints show noticeably lower errors, which is consistent with their lower articulation and stable geometry. In contrast, distal joints including the elbows, wrists, knees, and ankles exhibit higher errors due to increased articulation, more frequent occlusion, and greater appearance variability. The Pelvis joint shows the highest raw error value because it serves as the alignment root and therefore its magnitude reflects residual global offsets rather than local pose quality.

Overall, PoseAug, despite having never been trained on our synthetic data, successfully reconstructs coherent 3D skeletal structures from 2D keypoints. The error trends closely match observations in real-world motion capture datasets, providing strong evidence that our synthetic data maintain high geometric reliability and are suitable for downstream 2D → 3D lifting, domain adaptation, and cross-domain regression tasks.

\subsection{Image → 3D Joint Regression}
\begin{figure}[t]
\centering
\includegraphics[width=1\linewidth]{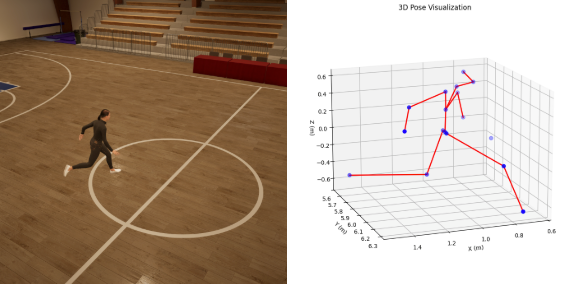}
\caption{Single-person synthetic image (left) and the corresponding 3D pose predicted by MeTRAbs (right).}
\label{fig:img_3d_single}
\end{figure}

\begin{figure}[t]
\centering
\includegraphics[width=1\linewidth]{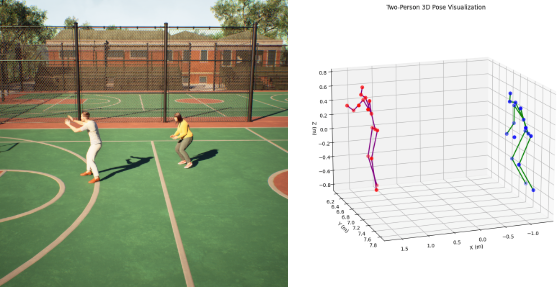}
\caption{Two-person synthetic image (left) and the corresponding 3D poses predicted by MeTRAbs (right).}
\label{fig:img_3d_two}
\end{figure}

To evaluate how well our synthetic data supports 3D human pose estimation, we use MeTRAbs (Absolute 3D Human Pose Estimator) \cite{sarandi2020metrabs} as an external inference model and compare its predicted 3D joints against the ground-truth annotations provided by our synthetic images. MeTRAbs is trained on multiple real-world datasets and has strong cross-scene generalization ability, yet it has never been fine-tuned on our synthetic data. This makes it a meaningful independent evaluator of the geometric consistency between our rendered images and their 3D annotations. Representative qualitative results are shown in Figure~\ref{fig:img_3d_single} (single-person) and Figure~\ref{fig:img_3d_two} (two-persons), where the left side shows the synthetic RGB input and the right side displays the inferred 3D pose(s).

\begin{figure*}[t]
    \centering
    \includegraphics[width=\textwidth]{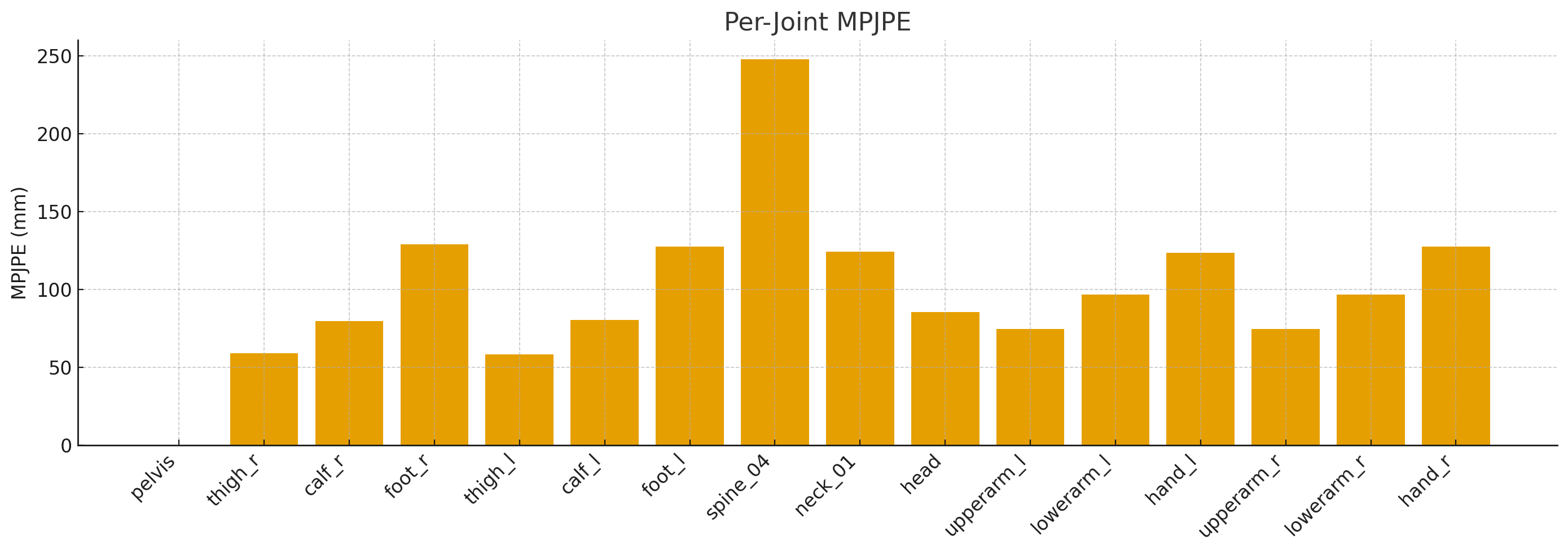}
    \caption{Per-joint MPJPE evaluation of MeTRAbs on synthetic Image-to-3D pose regression.}
    \label{fig:Per_joint_matrabs}
\end{figure*}

We evaluate our test set using MeTRAbs ~\cite{sarandi2020metrabs}. We predict 3D positions for the 17 Human3.6M joints and convert them into the 16-joint format of UnrealPose-1M through a predefined joint mapping. We apply root-joint translation alignment to ensure that the evaluation focuses on the relative body geometry rather than global position. To measure structural errors we report three metrics: MPJPE, PA-MPJPE, and per-joint MPJPE.

MeTRAbs achieves an MPJPE of 99.17 mm and a PA-MPJPE of 100.51 mm on approximately 38,000 sampled images, as shown in Table~\ref{tab:quant}. This error magnitude is consistent with its performance in other cross-domain scenarios, indicating strong correspondence between our synthetic images and real imagery. The per-joint results (Figure~\ref{fig:Per_joint_matrabs}) show lower errors for central body joints (such as Hip and other core torso joints), while higher errors are found in distal joints (including the neck, wrists, and ankles). This trend aligns with typical characteristics in cross-domain 3D pose estimation. The torso region exhibits stable texture and clear shape, resulting in minimal cross-domain impact. However, end joints are more susceptible to changes in viewpoint, occlusion, and differences in rendering details, resulting in typically higher cross-domain errors. It is noteworthy that the root joint error is 0, determined by the method of root alignment calculation and not used to assess pose quality.

Overall, MeTRAbs reconstructs structurally coherent 3D skeletons on our synthetic data without fine-tuning, with error patterns matching those on real-world datasets, validating the fidelity of UnrealPose-1M.

\subsection{Image → Person Instance Segmentation}
We evaluate person instance segmentation using Mask2Former ~\cite{m2f} for panoptic segmentation with a Swin-L backbone, pretrained on COCO ~\cite{lin2015microsoft}. The model achieves an average IoU of 0.89 on our test set, successfully segmenting individuals in multi-person scenarios with occlusions. Importantly, diverse scene elements --- including sky, vases, trees, etc --- are also correctly labeled, demonstrating the realism of our environments. This validates our MetaHuman rendering and ground-truth masks, as well as the overall scene realism necessary for synthetic data to serve as effective training material for real-world applications.

\begin{figure}[t]
\centering
\includegraphics[width=0.9\linewidth]{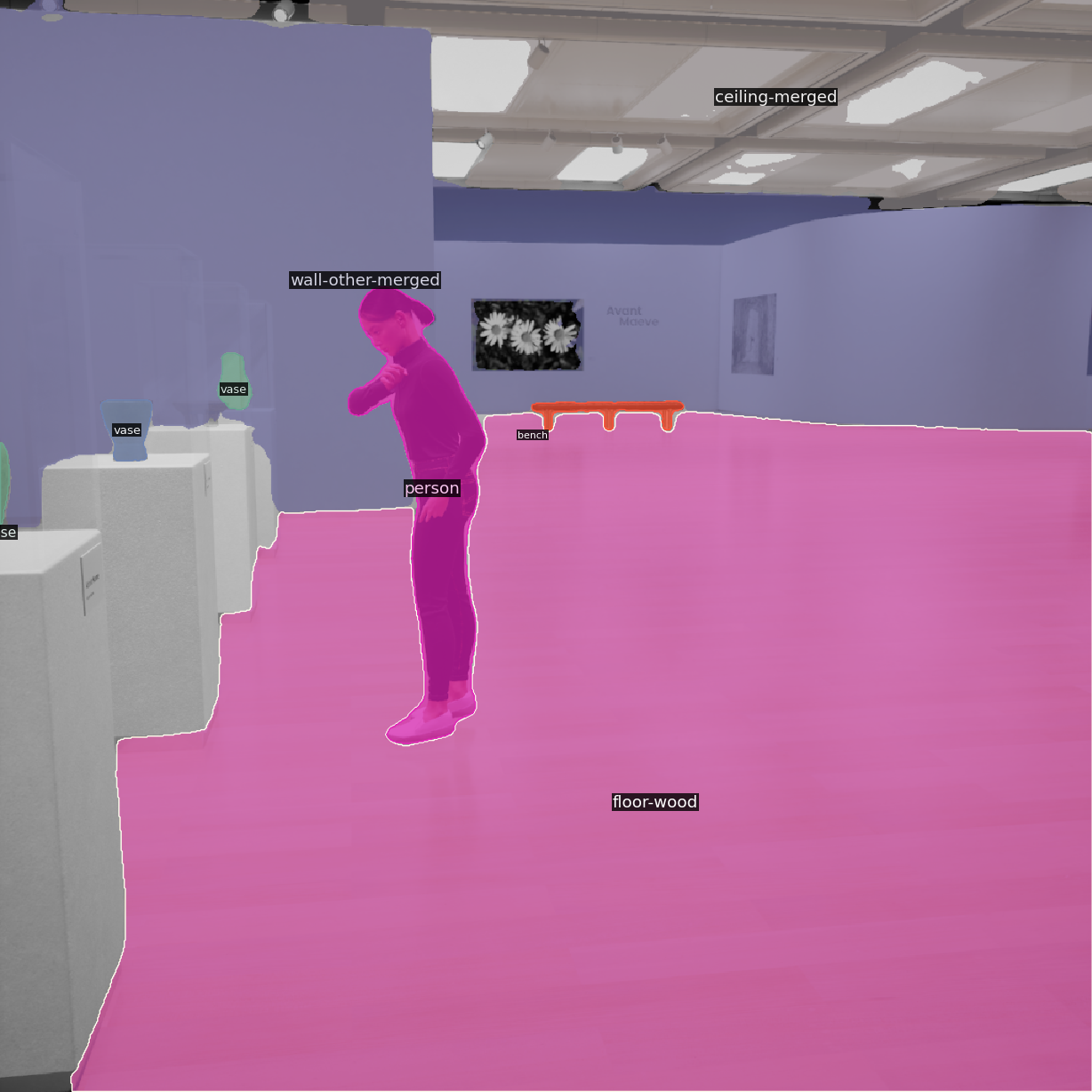}
\caption{Example panoptic segmentation results using Mask2Former on UnrealPose-1M.}
\label{fig:panoptic_seg}
\end{figure}

\section{Limitations and Future Work}
\label{sec:lfw}
\noindent\textbf{Pipeline Portability.}
UnrealPose-Gen currently requires manual integration into UE5 projects. An important next step is packaging the system as a UE5 plugin that can easily be integrated into any project. This would dramatically lower the barrier to entry, enabling researchers and developers to generate pose data without having to modify their project or understand our codebase.

\noindent\textbf{Scaling Characters, Skeletons, Animations, and Dataset Size.}
Our five MetaHumans represent a tiny fraction of the diverse body types possible with the MetaHuman creator, which can generate thousands of unique characters with varying characteristics. Furthermore, our pipeline's design allows it to work with any UE compatible mesh, not just MetaHumans. Future work should explore other character models, amputees, non-human entities, etc. The UE marketplace contains thousands of character models and animations. Systematically scaling up character diversity, skeleton types, and animation libraries would demonstrate the full generality of our approach. Additionally, while UnrealPose-1M contains one million frames, the pipeline is only limited by time and compute in generating datasets. Investigating the scaling laws of synthetic pose data remains an important question.

\noindent\textbf{Moving Cameras and Dynamic Intrinsics.}
A very simple change is to extend our pipeline to support moving cameras with varying intrinsics. This would enable the simulation of more realistic footage and potentially reduce the need for as many static camera positions.

\noindent\textbf{Compute Constraints and Model Training.}
Due to computational constraints, we were unable to perform training of pose estimation models on UnrealPose-1M. Our experiments focus on evaluation using pretrained models to validate data quality and fidelity. Future work will explore training models from scratch as well as fine-tuning on UnrealPose-1M to assess its potential for synthetic-to-real transfer and compare its performance to other synthetic datasets.

\noindent\textbf{Online Rendering in Real Games.}
Since UnrealPose-1M uses MRQ for maximum quality, we have not yet validated the online rendering capability in production game environments. Testing UnrealPose-Gen in UE5 games would demonstrate the practical utility of this capability.

\section{Conclusion}
\label{sec:dnc}
In this paper we present a UE-native approach to synthetic human pose data as opposed to the common body mesh approaches. UnrealPose-Gen is a UE5 pipeline for generating human pose data and UnrealPose-1M is a dataset demonstrating its capabilities. By leveraging game engine technology and content, we provide an approach to synthetic data generation that is more accessible, scalable, and rich in complex interactions. Our experiments demonstrate that even with a limited dataset we achieve competitive performance on benchmarks. We release both the pipeline and dataset publicly to support future research and encourage the community to explore game animation content for computer vision research.
{\small
\bibliographystyle{ieeenat_fullname}
\bibliography{main}
}
\end{document}